\documentclass{llncs}

\usepackage[utf8]{inputenc}
\usepackage{amssymb}
\usepackage{mathtools}
\usepackage{bussproofs}
\usepackage{stmaryrd}
\makeatletter
\def\new@fontshape{}
\makeatother
\usepackage{gb4e}
\noautomath
\usepackage{url}
\usepackage{hyperref}
\usepackage{subcaption}

\newcommand{\hsbind}{\mathbin{\gg\!=}}
\newcommand{\apl}{\mathbin{\ll\!\!\cdot}}
\newcommand{\apr}{\mathbin{\cdot\!\!\gg}}
\newcommand{\aplr}{\mathbin{\ll\!\!\cdot\!\!\gg}}

\newcommand{\abs}[1]{\textsc{#1}}
\newcommand{\obj}[1]{\textbf{#1}}
\newcommand{\sem}[1]{\llbracket #1 \rrbracket}
\newcommand{\lex}[2]{\sem{\abs{#1}} &:= #2}

\newcommand{\dand}{\mathbin{\overline{\land}}}

\newcommand{\dimpl}{\mathbin{\overline{\to}}}

\newcommand{\limp}{\mathbin{{-}\mkern-3.5mu{\circ}}}

\newcommand{\lban}{\llparenthesis \,}
\newcommand{\rban}{\, \rrparenthesis}

\newcommand{\banana}[1]{\lban #1 \rban}

\newcommand{\cherry}{\rotatebox[origin=c]{270}{$\limp$}}

\newcommand{\lam}[2]{\lambda #1.\, #2}
\newcommand{\ap}[2]{#1\,#2}
\newcommand{\app}[3]{\ap{\ap{#1}{#2}}{#3}}

\newcommand{\op}[1]{\mathtt{#1}}
\newcommand{\onto}[1]{#1 \mathalpha{:\,}}
\newcommand{\typedop}[3]{\op{#1} : #2 \rightarrowtail #3}
\newcommand{\typedopg}[3]{#1 : #2 \rightarrowtail #3}

\newcommand{\CC}{\mathcal{C}}
\newcommand{\FF}{\mathcal{F}}
\newcommand{\XX}{\mathcal{X}}
\newcommand{\EE}{\mathcal{E}}
\newcommand{\TT}{\mathcal{T}}

\newcommand{\FV}{\operatorname{FV}}

\newcommand{\subst}[3]{#1[#2 \coloneqq #3]}

\newcommand{\cibanana}{\banana{(\onto{\op{op}_i} M_i)_{i \in I},\ \onto{\eta} M_\eta}}
\newcommand{\cdbanana}{\banana{\onto{\op{op}_1} M_1,\ \dots,\ \onto{\op{op}_n} M_n,\ \onto{\eta} M_\eta}}

\newcommand{\tto}{\twoheadrightarrow}

\mathchardef\mhyphen="2D

\newcommand{\etaE}[1]{\ap{\eta}{#1}}

\newcommand{\withSpeaker}{\operatorname{withSpeaker}}
\newcommand{\SI}{\operatorname{SI}}
\newcommand{\accommodate}{\operatorname{accommodate}}

\newcommand{\negSpaceBeforeAlign}{\vspace{-4mm}}
\newcommand{\negSpaceBetweenAlign}{\vspace{-8mm}}

\begin{document}
\title{Introducing a Calculus of Effects and Handlers \\ for Natural Language Semantics}
\author{Jirka Maršík \and Maxime Amblard}

\institute{LORIA, UMR 7503, Université de Lorraine, CNRS, Inria, Campus Scientifique, \\
F-54506 Vand\oe uvre-lès-Nancy, France \\
\email{\{jirka.marsik, maxime.amblard\}@loria.fr}}

\titlerunning{Calculus of Effects and Handlers}
\authorrunning{Maršík and Amblard}

\maketitle              

\begin{abstract}
In compositional model-theoretic semantics, researchers assemble
truth-conditions or other kinds of denotations using the lambda
calculus. It was previously observed~\cite{shan2002monads} that the lambda
terms and/or the denotations studied tend to follow the same pattern: they
are instances of a monad. In this paper, we present an extension of the
simply-typed lambda calculus that exploits this uniformity using the
recently discovered technique of effect
handlers~\cite{plotkin2009handlers}. We prove that our calculus exhibits
some of the key formal properties of the lambda calculus and we use it to
construct a modular semantics for a small fragment that involves multiple
distinct semantic phenomena.

\end{abstract}

\section{Introduction}
\label{sec:introduction}

The prevailing methodology of formal semantics is compositionality in the
sense of Frege: denotations of complex phrases are functions of the
denotations of their immediate constituents. However, several phenomena
have been identified that challenge this notion of
compositionality. Examples include anaphora, presupposition,
quantification, deixis and conventional implicature. In all of these
examples, simple models of denotation (i.e.\ noun phrases are individuals,
sentences are truth-values) run into complications as the denotations can
depend on external values (anaphora, deixis) or on something which is not
an immediate constituent (presupposition, quantification, conventional
implicature).

Among the solutions to these challenges, we find (at least) two types of
solutions. First, we have those that relax the condition of
compositionality. Notably, the denotation of a complex phrase is no longer
a \emph{function per se} of the denotations of its immediate
subconstituents. Rather, it is some other formally defined
process.\footnote{This kind of distinction is the same distinction as the
  one between a mathematical function and a function in a programming
  language, which might have all kinds of side effects and therefore not be
  an actual function.} Examples of this approach include:

\begin{itemize}
\item the incremental algorithm used to build discourse representation
  structures in DRT, as presented in~\cite{kamp1993discourse}
\item the $\lambda\mu$ calculus, used in~\cite{de2001type} to analyze
  quantification, since, due to the lack of confluence, function terms do
  not denote functions over simple denotations
\item the use of exceptions and exception handlers
  in~\cite{lebedeva2012expression} to model presuppositions in an otherwise
  compositional framework
\item the parsetree interpretation step in the logic of conventional
  implicatures of~\cite{potts2005logic} that builds the denotation of a
  sentence by extracting implicatures from the denotations of its subparts
  (including the non-immediate ones)
\end{itemize}

The other approach is to enrich the denotations so that they are
parameterized by the external information they need to obtain and contain
whatever internal information they need to provide to their
superconstituents. Here are some examples of this style:

\begin{itemize}
\item any kind of semantic indices (e.g.\ the speaker and addressee for
  deixis, the current world for modality), since they amount to saying that
  a phrase denotes an indexed set of simpler meanings
\item the continuized semantics for
  quantification~\cite{barker2002continuations} in which denotations are
  functions of their own continuations
  \begin{itemize}
  \item and more generally, any semantics using type raising or generalized
    quantifiers for noun phrase denotations
  \end{itemize}
\item the dynamic denotations of~\cite{de2006towards} that are functions of
  the common ground and their continuation
\item compositional event semantics, such as the one
  in~\cite{qian2011event}, that shift the denotations of sentences from
  truth-values to predicates on events
\end{itemize}

We want to find a common language in which we could express the above
techniques. Our inspiration comes from computer science. There, a concept
known as \emph{monad} has been used:
\begin{itemize}
\item in denotational semantics to give the domain of interpretation for
  programming languages that involve side effects~\cite{moggi1991notions}.
\item in functional programming to emulate programming with side effects
  via term-level encodings of effectful programs~\cite{wadler1992essence}.
\end{itemize}
These two principal applications of monads align with the two approaches we
have seen above. The one where we change our calculus so it no longer
defines pure functions (e.g.\ is non-deterministic, stateful or throws
exceptions) and the one where we use a pure calculus to manipulate terms
(denotations) that encode some interaction (e.g.\ dynamicity, continuations
or event predication).

Monad is a term from category-theory. Its meaning is relative to a
category. For us, this will always be the category whose objects are types
and whose arrows are functions between different types. A monad is formed
by a functor and a pair of natural transformations that satisfy certain
laws. In our case, this means that a monad is some type constructor (the
functor part) and some combinators (the natural transformations) that
follow some basic laws. To give an example of this, we can think of the
functor $T(\alpha) = (\alpha \to o) \to o$ together with combinators such
as the type raising $\eta(x) = \lam{P}{\ap{P}{x}}$ as a monad of
quantification.

The relationship between side effects in functional programming and
computational semantics has been developed in several works
\cite{shan2005thesis,van2010computational},\footnote{Side effects are
  to programming languages what pragmatics are to natural languages: they
  both study how expressions interact with the worlds of their users. It
  might then come as no surprise that phenomena such as anaphora,
  presupposition, deixis and conventional implicature yield a monadic
  description.} stretching as far back as 1977~\cite{hobbs1977making}. The
usefulness of monads in particular has been discovered by Shan in
2002~\cite{shan2002monads}. Since then, the problem that remained was how
to compose several different monads in a single solution. Charlow used the
popular method of monad morphisms\footnote{Also known as monad transformers
  in functional programming.} to combine several monads in his
dissertation~\cite{charlow2014semantics}. Giorgolo and Asudeh have used
distributive laws to combine monads~\cite{giorgolo2015natural}, while
Kiselyov has eschewed monads altogether in favor of applicative functors
which enjoy easy composability~\cite{kiselyov2015applicative}.

Our approach follows the recent trend in adopting effects and handlers to
combine side effects~\cite{bauer2012programming,kammar2013handlers} and to
encode effectful programs in pure functional programming
languages~\cite{kiselyov2013extensible,brady2013programming}.

The idea is that we can represent each of the relevant monads using an
algebra. We can then combine the signatures of the algebras by taking a
disjoint union. The free algebra of the resulting signature will serve as a
universal representation format for the set of all terms built from any of
the source algebras and closed under substitution. Then, we will build
modular interpreters that will give meanings to the operators of the
algebras in terms of individuals, truth-values and functions.

In Sect.~\ref{sec:definition}, we will introduce a formal calculus for
working with the algebraic terms that we will use in our linguistic
denotations. In Sect.~\ref{sec:phenomena}, we will incrementally build up a
fragment involving several of the linguistic phenomena and see the calculus
in action. Before we conclude in Sect.~\ref{sec:conclusion}, we will also
discuss some of the formal properties of the calculus in
Sect.~\ref{sec:properties}.

\section{Definition of the Calculus}
\label{sec:definition}

Our calculus is an extension of the simply-typed lambda calculus (STLC). We
add terms of a free algebra into our language and a notation for writing
handlers, composable interpreters of these terms. An operator of the free
algebra corresponds to a particular interaction that a piece of natural
language can have with its context (e.g.\ a deictic expression might
request the speaker's identity using some operator $\op{speaker}$ in order
to find its denotation). A handler gives an interpretation to every
occurrence of an operator within a term (e.g.\ direct speech introduces a
handler for the operator $\op{speaker}$ that essentially rebinds the
current speaker to some other entity).

Having sketched the general idea behind our calculus, we will now turn our
attention to the specifics. We start by defining the syntactic
constructions used to build the terms of our language.

\subsection{Terms}
\label{ssec:terms}

First off, let $\XX$ be a set of variables, $\Sigma$ a typed signature and
$\EE$ a set of operation symbols. In the definition below, we will let $M$,
$N$\ldots range over terms,\, $x$, $y$, $z$\ldots range over variables from
$\XX$,\, $c$, $d$\ldots range over the names of constants from $\Sigma$ and
$\op{op}$, $\op{op}_i$\ldots range over the operation symbols in $\EE$.

The terms of our language are composed of the following:

\begin{align*}
  M, N ::= &\ \lam{x}{M} & \mbox{[abstraction]} \\
   | \, &\ \ap{M}{N} & \mbox{[application]} \\
   | \, &\ x & \mbox{[variable]} \\
   | \, &\ c & \mbox{[constant]} \\
   | \, &\ \app{\op{op}}{M_{\mathrm{p}}}{(\lam{x}{M_{\mathrm{c}}})} & \mbox{[operation]} \\
   | \, &\ \ap{\eta}{M} & \mbox{[injection]} \\
   | \, &\ \ap{\cdbanana}{N} & \mbox{[handler]} \\
   | \, &\ \ap{\cherry}{M} & \mbox{[extraction]} \\
   | \, &\ \ap{\CC}{M} & \mbox{[exchange]} 
\end{align*}

The first four constructions --- abstraction, application, variables and
constants --- come directly from STLC with constants.

The next four deal with the algebraic expressions used to encode
computations. Let us sketch the behaviors of these four kinds of
expressions.

The operation ($\op{op}$) and injection ($\eta$) expressions will serve as
the constructors for our algebraic expressions. Algebraic expressions are
usually formed by operation symbols and then variables as atoms. Instead of
variables, our algebraic expressions use terms from our calculus for
atoms. The $\eta$ constructor can thus take an ordinary term from our
calculus and make it an atomic algebraic expression. The operation symbols
$\op{op}$ are then the operations of the algebra.

The other three expression types correspond to functions over algebraic
expressions.
\begin{itemize}
\item The most useful is the handler $\banana{}$.\footnote{Pronounced
  ``banana''. See~\cite{meijer1991functional} for the introduction of
  banana brackets.} It is an iterator for the type of algebraic
  expressions. The terms $M_1$,\ldots,$M_n$ and $M_\eta$ in $\cdbanana$ are
  the clauses for the constructors $\op{op}_1$,\ldots,$\op{op}_n$ and
  $\eta$, respectively. We will use handlers to define interpretations of
  operation symbols in algebraic expressions.
\item The cherry $\cherry$ operator allows us to extract terms out of
  algebraic expressions. If an algebraic expression is of the form
  $\ap{\eta}{M}$, applying $\cherry$ to it will yield $M$.
\item The exchange operator $\CC$ permits a kind of commutation between the
  $\lambda$-binder and the operation symbols. We will see its use later.
\end{itemize}

\subsection{Types}
\label{ssec:types}

We now give a syntax for the types of our calculus along with a typing
relation. In the grammar below,\, $\alpha$, $\beta$, $\gamma$\ldots range
over types,\, $\nu$ ranges over atomic types from some set $\TT$ and $E$,
$E'$\ldots range over effect signatures (introduced below).

The types of our language consist of:

\negSpaceBeforeAlign

\begin{align*}
  \alpha, \beta, \gamma ::= &\ \alpha \to \beta & \mbox{[function]} \\
   | \, &\ \nu & \mbox{[atom]} \\
   | \, &\ \FF_E(\alpha) & \mbox{[computation]}
\end{align*}

The only novelty here is the $\FF_E(\alpha)$ computation type. This is the
type of algebraic expressions whose atoms are terms of type $\alpha$ and
whose operation symbols come from the effect signature $E$. We call them
\emph{computation types} and we call terms of these types
\emph{computations} because our algebraic expressions will always represent
some kind of program with effects.

\emph{Effect signatures} are similar to typing contexts. They are partial
mappings from the set of operation symbols $\EE$ to pairs of types. We will
write the elements of effect signatures the following way ---
$\typedop{op}{\alpha}{\beta} \in E$ means that $E$ maps $\op{op}$ to the
pair of types $\alpha$ and $\beta$.\footnote{The two types $\alpha$ and
  $\beta$ are to be seen as the operation's \emph{input} and \emph{output}
  types, respectively.} When dealing with effect signatures, we will often
make use of the disjoint union operator $\uplus$. The term $E_1 \uplus E_2$
serves as a constraint demanding that the domains of $E_1$ and $E_2$ be
disjoint and at the same time it denotes the effect signature that is the
union of $E_1$ and $E_2$.

The typing rules are presented in Figure~\ref{fig:types}.

\newcommand{\handlerrule}{
 \begin{prooftree}
  \AxiomC{$E = \{\typedopg{\op{op}_i}{\alpha_i}{\beta_i}\}_{i \in I} \uplus E_{\mathrm{f}}$}
  \noLine
  \def\extraVskip{0pt}
  \UnaryInfC{$E' = E'' \uplus E_{\mathrm{f}}$}
  \noLine
  \UnaryInfC{$[\Gamma \vdash M_i : \alpha_i \to (\beta_i \to
    \FF_{E'}(\delta)) \to \FF_{E'}(\delta)]_{i \in I}$}
  \noLine
  \UnaryInfC{$\Gamma \vdash M_\eta : \gamma \to \FF_{E'}(\delta)$}
  \noLine
  \UnaryInfC{$\Gamma \vdash N : \FF_{E}(\gamma)$}
  \def\extraVskip{2pt}
  \RightLabel{[$\banana{}$]}
  \UnaryInfC{$\Gamma \vdash \ap{\cibanana}{N} : \FF_{E'}(\delta)$}
 \end{prooftree}}

\begin{figure}
  \def\labelSpacing{4pt}

  \begin{subfigure}{.5\textwidth}
   \begin{prooftree}
    \AxiomC{$\Gamma, x : \alpha \vdash M : \beta$}
    \RightLabel{[abs]}
    \UnaryInfC{$\Gamma \vdash \lam{x}{M} : \alpha \to \beta$}
   \end{prooftree}
  \end{subfigure}
  \begin{subfigure}{.5\textwidth}
   \begin{prooftree}
    \AxiomC{$\Gamma \vdash M : \alpha \to \beta$}
    \AxiomC{$\Gamma \vdash N : \alpha$}
    \RightLabel{[app]}
    \BinaryInfC{$\Gamma \vdash M N : \beta$}
   \end{prooftree}
  \end{subfigure}

  \vspace{2mm}
 
  \begin{subfigure}{.5\textwidth}
   \begin{prooftree}
    \AxiomC{$x : \alpha \in \Gamma$}
    \RightLabel{[var]}
    \UnaryInfC{$\Gamma \vdash x : \alpha$}
   \end{prooftree}
  \end{subfigure}
  \begin{subfigure}{.5\textwidth}
   \begin{prooftree}
    \AxiomC{$c : \alpha \in \Sigma$}
    \RightLabel{[const]}
    \UnaryInfC{$\Gamma \vdash c : \alpha$}
   \end{prooftree}
  \end{subfigure}

  \vspace{6mm}

  \begin{subfigure}{.5\textwidth}
   \begin{prooftree}
    \AxiomC{$\Gamma \vdash M : \alpha$}
    \RightLabel{[$\eta$]}
    \UnaryInfC{$\Gamma \vdash \ap{\eta}{M} : \FF_E(\alpha)$}
   \end{prooftree}
  \end{subfigure}
  \begin{subfigure}{.5\textwidth}
   \begin{prooftree}
    \AxiomC{$\Gamma \vdash M_{\mathrm{p}} : \alpha$}
    \AxiomC{$\Gamma, x : \beta \vdash M_{\mathrm{c}} : \FF_E(\gamma)$}
    \def\extraVskip{0pt}
    \noLine
    \BinaryInfC{$\typedop{op}{\alpha}{\beta} \in E$}
    \def\extraVskip{2pt}
    \RightLabel{[op]}
    \UnaryInfC{$\Gamma \vdash \app{\op{op}}{M_{\mathrm{p}}}{(\lam{x}{M_{\mathrm{c}}})} : \FF_E(\gamma)$}
   \end{prooftree}
  \end{subfigure}

  \vspace{3mm}

  \hspace{-1.5cm}
  \begin{subfigure}{.5\textwidth}
   \begin{prooftree}
    \AxiomC{$\Gamma \vdash M : \FF_\emptyset(\alpha)$}
    \RightLabel{[$\cherry$]}
    \UnaryInfC{$\Gamma \vdash \ap{\cherry}{M} : \alpha$}
   \end{prooftree}
  \end{subfigure}
  \hspace{1cm}
  \begin{subfigure}{.5\textwidth}
   \handlerrule
  \end{subfigure}

  \vspace{6mm}

  \begin{subfigure}{\textwidth}
   \begin{prooftree}
    \AxiomC{$\Gamma \vdash M : \alpha \to \FF_E(\beta)$}
    \RightLabel{[$\CC$]}
    \UnaryInfC{$\Gamma \vdash \ap{\CC}{M} : \FF_E(\alpha \to \beta)$}
   \end{prooftree}
  \end{subfigure}

  \caption{\label{fig:types}The typing rules for our calculus.}
\end{figure}

The typing rules mirror the syntax of terms. Again, the first four rules
come from STLC.\@ The [$\eta$] and [$\cherry$] rules are self-explanatory
and so we will focus on the [$\op{op}$], [$\banana{}$] and [$\CC$] rules.

\subsubsection*{[$\op{op}$]}

To use an operation $\typedop{op}{\alpha}{\beta}$, we provide the input
parameter $M_{\mathrm{p}} : \alpha$ and a continuation
$\lam{x}{M_{\mathrm{c}}} : \beta \to \FF_E(\gamma)$, which expects the
output of type $\beta$. The resulting term has the same type as the body of
the continuation, $\FF_E(\gamma)$.

Before, we have spoken of terms of type $\FF_E(\gamma)$ as of algebraic
expressions generated by the terms of type $\gamma$ and the operators in
the effect signature $E$. However, having seen the typing rule for
operation terms, it might not be obvious how such a term represents an
algebraic expression. Traditionally, algebraic signatures map operation
symbols to arities, which are natural numbers. Our effect signatures map
each operation symbol to a pair of types $\alpha \rightarrowtail \beta$.
\begin{itemize}
\item We can explain $\alpha$ by analogy to the single-sorted algebra of
  vector spaces. In a single-sorted vector space algebra, scalar
  multiplication is viewed as a unary operation parameterized by some
  scalar. So technically, there is a different unary operation for each
  scalar. All of our operations are similarly parameterized and $\alpha$ is
  the type of that parameter.
\item The type $\beta$ expresses the arity of the operator. When we say
  that an operator has arity $\beta$, where $\beta$ is a type, we mean that
  it takes one operand for every value of $\beta$
  \cite{pretnar2010logic}. We can also think of the operator as taking one
  operand containing $x : \beta$ as a free variable.
\end{itemize}

We can look at the algebraic expression
$\app{\op{op}}{M_{\mathrm{p}}}{(\lam{x}{M_{\mathrm{c}}})}$ as a description
of a program that:
\begin{itemize}
\item interacts with its context by some operator called $\op{op}$
\item to which it provides the input $M_{\mathrm{p}}$
\item and from which it expects to receive an output of type $\beta$
\item which it will then bind as the variable $x$ and continue as the
  program described by $M_{\mathrm{c}}$.
\end{itemize}

\subsubsection*{[$\banana{}$]}

The banana brackets describe iterators/catamorphisms.\footnote{These are
  similar to recursors/paramorphisms. See~\cite{meijer1991functional} for
  the difference. Catamorphisms are also known as folds and the common
  higher-order function \emph{fold} found in functional programming
  languages is actually the iterator/catamorphism for lists.} In the typing
rule, $E$ is the input's signature, $E'$ is the output's signature,
$\gamma$ is the input's atom type and $\delta$ is the output's atom
type. $E$ is decomposed into the operations that our iterator will actually
interpret, the other operations form a residual signature
$E_{\mathrm{f}}$. The output signature will then still contain the
uninterpreted operations $E_{\mathrm{f}}$ combined with any operations
$E''$ that our interpretation might introduce.

\subsubsection*{[$\CC$]}

We said before that the $\CC$ function will let us commute $\lambda$ and
operations. Here we see that, on the type level, this corresponds to
commuting the $\FF_E(\_)$ and the $\alpha \to \_$ type constructors.

\subsection{Reduction Rules}
\label{ssec:reductions}

We will now finally give a semantics to our calculus. The semantics will be
given in the form of a reduction relation on terms. Even though the point
of the calculus is to talk about effects, the reduction semantics will not
be based on any fixed evaluation order; any subterm that is a redex can be
reduced in any context. The reduction rules are given in
Fig.~\ref{fig:reductions}.

\begin{figure}
  \centering
  \begin{tabular}{lr}
  $\ap{(\lam{x}{M})}{N} \to$ & rule $\beta$ \\
  $\subst{M}{x}{N}$ & \\
  \\
  $\lam{x}{\ap{M}{x}} \to$ & rule $\eta$ \\
  $M$ & where $x \notin \FV(M)$ \\
  \\
  $\ap{\cibanana}{(\ap{\eta}{N})} \to$ & rule $\banana{\eta}$ \\
  $\ap{M_\eta}{N}$ & \\
  \\
  $\ap{\cibanana}{(\ap{\ap{\op{op}_j}{N_{\mathrm{p}}}}{(\lam{x}{N_{\mathrm{c}}})})} \to$ & rule $\banana{\op{op}}$ \\
  $\ap{M_j}{\ap{N_{\mathrm{p}}}{(\lam{x}{\ap{\cibanana}{N_{\mathrm{c}}}})}}$
  & where $j \in I$ \\
  & and $x \notin \FV((M_i)_{i \in I}, M_\eta)$ \\
  \\
  $\ap{\cibanana}{(\ap{\ap{\op{op}_j}{N_{\mathrm{p}}}}{(\lam{x}{N_{\mathrm{c}}})})} \to$ & rule $\banana{\op{op}'}$ \\
  $\ap{\op{op}_j}{\ap{N_{\mathrm{p}}}{(\lam{x}{\ap{\cibanana}{N_{\mathrm{c}}}})}}$
  & where $j \notin I$ \\
  & and $x \notin \FV((M_i)_{i \in I}, M_\eta)$ \\
  \\
  $\ap{\cherry}{(\ap{\eta}{M})} \to$ & rule $\cherry$ \\
  $M$ & \\
  \\
  $\ap{\CC}{(\lam{x}{\ap{\eta}{M}})} \to$ & rule $\CC_\eta$ \\
  $\ap{\eta}{(\lam{x}{M})}$ & \\
  \\
  $\ap{\CC}{(\lam{x}{\ap{\ap{\op{op}}{M_{\mathrm{p}}}}{(\lam{y}{M_{\mathrm{c}}})}})} \to$ & rule $\CC_\op{op}$ \\
  $\ap{\ap{\op{op}}{M_{\mathrm{p}}}}{(\lam{y}{\ap{\CC}{(\lam{x}{M_{\mathrm{c}}})}})}$
  & where $x \notin \FV(M_{\mathrm{p}})$
  \end{tabular}
  
  \caption{\label{fig:reductions} The reduction rules of our calculus.}
\end{figure}

We have the $\beta$ and $\eta$ rules, which, by no coincidence, are the
same rules as the ones found in STLC.\@ The rest are function definitions
for $\banana{}$, $\cherry$ and $\CC$.

By looking at the definition of $\banana{}$, we see that it is an
iterator. It replaces every occurrence of the constructors $\op{op}_j$ and
$\eta$ with $M_j$ and $M_\eta$, respectively.

The $\CC$ function recursively swaps $\ap{\CC}{(\lam{x}{\_})}$ with
$\app{\op{op}}{M_{\mathrm{p}}}{(\lam{y}{\_})}$ using the $\CC_{\op{op}}$
rule. When $\CC$ finally meets the $\eta$ constructor, it swaps
$(\lam{x}{\_})$ with $\ap{\eta}{\_}$ and terminates. Note that the
constraint $x \notin \FV(M_{\mathrm{p}})$ in rule $\CC_{\op{op}}$ cannot be
dismissed by renaming of bound variables. If the parameter $M_{\mathrm{p}}$
contains a free occurrence of $x$, the evaluation of $\CC$ will get
stuck. $\CC$ is thus a partial function: it is only applicable when none of
the operations being commuted with the $\lambda$-binder actually depend on
the bound variable.

\subsection{Common Combinators}
\label{ssec:common-combinators}

When demonstrating the calculus in the next section, the following
combinators will be helpful. First, we define a sequencing operator. The
operator $\hsbind$, called bind, replaces all the $\alpha$-typed atoms of a
$\FF_E(\alpha)$-typed expression with $\FF_E(\beta)$-typed
expressions. More intuitively, $M \hsbind N$ is the program that first runs
$M$ to get its result $x$ and then continues as the program $\ap{N}{x}$.

\negSpaceBeforeAlign

\begin{align*}
  & \_ \hsbind \_ : \FF_E(\alpha) \to (\alpha \to \FF_E(\beta)) \to \FF_E(\beta) \\
  & M \hsbind N = \ap{\banana{\onto{\eta}{N}}}{M}
\end{align*}

The type constructor $\FF_E$ along with the operators $\eta$ and $\hsbind$
form a free monad. Using this monadic structure, we can define the
following combinators (variations on application) which we will make heavy
use of in Section~\ref{sec:phenomena}.

\negSpaceBeforeAlign

\begin{align*}
  & \_ \apl \_ : \FF_E(\alpha \to \beta) \to \alpha \to \FF_E(\beta) \\
  & F \apl x = F \hsbind (\lam{f}{\ap{\eta}{(\ap{f}{x})}}) \\[0.8ex]
  & \_ \apr \_ : (\alpha \to \beta) \to \FF_E(\alpha) \to \FF_E(\beta) \\
  & f \apr X = X \hsbind (\lam{x}{\ap{\eta}{(\ap{f}{x})}}) \\[0.8ex]
  & \_ \aplr \_ : \FF_E(\alpha \to \beta) \to \FF_E(\alpha) \to \FF_E(\beta) \\
  & F \aplr X = F \hsbind (\lam{f}{X \hsbind (\lam{x}{\ap{\eta}{(\ap{f}{x})}})})
\end{align*}

All of these operators associate to the left, so $f \apr X \aplr Y$ should
be read as $(f \apr X) \aplr Y$.

Let $\circ : o \to o \to o$ be a binary operator on propositions. We define
the following syntax for the same operator lifted to computations of
propositions.

\negSpaceBeforeAlign

\begin{align*}
  & \_ \mathop{\overline{\circ}} \_ : \FF_E(o) \to \FF_E(o) \to \FF_E(o) \\
  & M \mathop{\overline{\circ}} N = (\lam{m n}{m \circ n}) \apr M \aplr N
\end{align*}

\section{Linguistic Phenomena as Effects}
\label{sec:phenomena}

\subsection{Deixis}
\label{ssec:deixis}

We will now try to use this calculus to do some semantics. Here is our
tectogrammar in an abstract categorial grammar
presentation~\cite{de2001towards}.

\negSpaceBeforeAlign

\begin{align*}
  \abs{John}, \abs{Mary}, \abs{me} &: NP \\
  \abs{loves} &: NP \limp NP \limp S
\end{align*}

And here is our semantics.

\negSpaceBeforeAlign

\begin{align*}
  \lex{John}{\etaE{\obj{j}}} \\
  \lex{Mary}{\etaE{\obj{m}}} \\
  \lex{me}{\app{\op{speaker}}{\star}{(\lam{x}{\etaE{x}})}} \\
  \lex{loves}{\lam{O S}{{\obj{love}} \apr S \aplr O}}
\end{align*}

In the semantics for $\sem{\abs{me}}$, we use the $\op{speaker}$ operation
to retrieve the current speaker and make it available as the value of the
variable $x$. The star~($\star$) passed to $\op{speaker}$ is a dummy value
of the unit type $1$.

This, and all the semantics we will see in this paper, satisfies a
homomorphism condition that whenever $M : \tau$, then $\sem{M} :
\sem{\tau}$. In our case, $\sem{NP} = \FF_E(\iota)$ and $\sem{S} =
\FF_E(o)$, where $\iota$ and $o$ are the types of individuals and
propositions, respectively. Of $E$, we assume that
$\typedop{speaker}{1}{\iota} \in E$, since that is the type of
$\op{speaker}$ used in our semantics.\footnote{$1$ is the unit type whose
  only element is written as $\star$.}

With this fragment, we can give meanings to trivial sentences like:

\begin{exe}
  \ex John loves Mary. \label{ex:trivial}
  \ex Mary loves me. \label{ex:deixis}
\end{exe}

whose meanings we can calculate as:

\negSpaceBeforeAlign

\begin{align}
  \sem{\app{\abs{loves}}{\abs{Mary}}{\abs{John}}} & \tto 
  \etaE{(\app{\obj{love}}{\obj{j}}{\obj{m}})} \\
  \sem{\app{\abs{loves}}{\abs{me}}{\abs{Mary}}} & \tto
  \app{\op{speaker}}{\star}{(\lam{x}{\etaE{(\app{\obj{love}}{\obj{m}}{x})}})}
\end{align}

The meaning of~\eqref{ex:trivial} is a proposition of type $o$ wrapped in
$\eta$, i.e.\ something that we can interpret in a model. As for the
meaning of~\eqref{ex:deixis}, the $\op{speaker}$ operator has propagated
from the $\abs{me}$ lexical entry up to the meaning of the whole
sentence. We now have an algebraic expression having as operands the
propositions $\app{\obj{love}}{\obj{m}}{x}$ for all possible $x :
\iota$. In order to get a single proposition which is to be seen as the
truth-conditional meaning of the sentence and which can be evaluated in a
model, we will need to fix the speaker. We will do so by defining an
interpreting handler.

\negSpaceBeforeAlign

\begin{align*}
  \withSpeaker &: \iota \to \FF_{\{\typedop{speaker}{1}{\iota}\} \uplus
    E}(\alpha) \to \FF_E(\alpha) \\
  \withSpeaker &= \lam{s M}{\ap{\banana{\onto{\op{speaker}}{(\lam{x k}{\ap{k}{s}})}}}{M}}
\end{align*}

Note that we omitted the $\eta$ clause in the banana brackets above. In
such cases, we say there is a default clause
$\onto{\eta}{(\lam{x}{\etaE{x}})}$.

$$
  \app{\withSpeaker}{s}{\sem{\app{\abs{loves}}{\abs{me}}{\abs{Mary}}}} \tto
  \etaE{(\app{\obj{love}}{\obj{m}}{s})}
$$

So far, we could have done the same by introducing a constant named
$\obj{me}$ to stand in for the speaker. However, since handlers are part of
our object language, we can include them in lexical entries. With this, we
can handle phenomena such as direct (quoted) speech, that rebinds the
current speaker in a certain scope.

\negSpaceBeforeAlign

\begin{align*}
  \abs{said}_{\abs{is}} &: S \limp NP \limp S \\
  \abs{said}_{\abs{ds}} &: S \limp NP \limp S
\end{align*}

Those are our new syntactic constructors: one for the indirect speech use
of \emph{said} and the other for the direct speech use (their surface
realizations would differ typographically or phonologically). Let us give
them some semantics.

\negSpaceBeforeAlign

\begin{align*}
  \sem{\abs{said}_{\abs{is}}} &= \lam{C S}{\obj{say} \apr S \aplr C} \\
                             &= \lam{C S}{S \hsbind (\lam{s}{\ap{\obj{say}}{s} \apr C})} \\
  \sem{\abs{said}_{\abs{ds}}} &= \lam{C S}{S \hsbind (\lam{s}{\ap{\obj{say}}{s} \apr (\app{\withSpeaker}{s}{C})})}
\end{align*}

Here we elaborated the entry for indirect speech so it is easier to compare
with the one for direct speech, which just adds a use of the $\withSpeaker$
operator.

\begin{exe}
  \ex John said Mary loves me. \label{ex:indirect-speech}
  \ex John said, ``Mary loves me''. \label{ex:direct-speech}
\end{exe}

\negSpaceBetweenAlign

\begin{align}
  \sem{\app{\abs{said}_{\abs{is}}}{(\app{\abs{loves}}{\abs{me}}{\abs{Mary}})}{\abs{John}}}
  & \tto \app{\op{speaker}}{\star}{(\lam{x}{\etaE{(\app{\obj{say}}{\obj{j}}{(\app{\obj{love}}{\obj{m}}{x})})}})} \\
  \sem{\app{\abs{said}_{\abs{ds}}}{(\app{\abs{loves}}{\abs{me}}{\abs{Mary}})}{\abs{John}}}
  & \tto \etaE{(\app{\obj{say}}{\obj{j}}{(\app{\obj{love}}{\obj{m}}{\obj{j}})})}
\end{align}

The meaning of sentence~\eqref{ex:indirect-speech} depends on the speaker
(as testified by the use of the $\op{speaker}$ operator) whereas
in~\eqref{ex:direct-speech}, this dependence has been eliminated due to the
use of direct speech.

\subsection{Quantification}
\label{ssec:quantification}

Now we turn our attention to quantificational noun phrases.

\negSpaceBeforeAlign

\begin{align*}
  \abs{every}, \abs{a} &: N \limp NP \\
  \abs{man}, \abs{woman} &: N
\end{align*}

\negSpaceBetweenAlign

\begin{align*}
  \lex{every}{\lam{N}{\app{\op{scope}}{(\lam{c}{\forall \apr
          (\ap{\CC}{(\lam{x}{(N \apl x) \dimpl (\ap{c}{x})})})})}{(\lam{x}{\etaE{x}})}}} \\
  \lex{a}{\lam{N}{\app{\op{scope}}{(\lam{c}{\exists \apr
          (\ap{\CC}{(\lam{x}{(N \apl x) \dand (\ap{c}{x})})})})}{(\lam{x}{\etaE{x}})}}} \\
  \lex{man}{\etaE{\obj{man}}} \\
  \lex{woman}{\etaE{\obj{woman}}}
\end{align*}

The entries for $\abs{every}$ and $\abs{a}$ might seem
intimidating. However, if we ignore the $\apr$, the $\CC$, the $\apl$ and
the overline on the logical operator, we get the familiar generalized
quantifiers. These decorations are the plumbing that takes care of the
proper sequencing of effects.

Note that we make use of the $\CC$ operator here. In the denotation of
$\sem{\abs{a}}$, the term $(\lam{x}{(N \apl x) \dand (\ap{c}{x})})$
describes the property to which we want to apply the quantifier
$\exists$. However, this term is of type $\iota \to \FF_E(o)$. In order to
apply $\exists$, we need something of type $\iota \to o$. Intuitively, the
effects of $E$ correspond to the process of interpretation, the process of
arriving at some logical form of the sentence. They should thus be
independent of the particular individual that we use as a witness for $x$
when we try to model-check the resulting logical form. This independence
allows us use the $\CC$ operator without fear of getting stuck. Once we
arrive at the type $\FF_E(\iota \to o)$, it is a simple case of using
$\exists \apr \_$ to apply the quantifier within the computation
type.\footnote{Other solutions to this problem include separating the
  language of logical forms and the metalanguage used in the semantic
  lexical entries to manipulate logical forms as
  objects~\cite{kiselyov2015applicative}.}\footnote{Our $\CC$ has been
  inspired by an operator of the same name proposed
  in~\cite{de2015conservativity}: de Groote introduces a structure that
  specializes applicative functors in a similar direction as monads by
  introducing the $\CC$ operator and equipping it with certain laws; our
  $\CC$ operator makes the $\FF_E$ type constructor an instance of this
  structure.}

While the terms that use the $\op{scope}$ operator might be complex, the
handler that interprets them is as simple as can be.

\negSpaceBeforeAlign

\begin{align*}
  \SI &= \lam{M}{\ap{\banana{\onto{\op{scope}}{(\lam{c k}{\ap{c}{k}})}}}{M}}
\end{align*}

Same as with $\withSpeaker$, $\SI$ will also be used in lexical items. By
interpreting the $\op{scope}$ operation in a particular place, we
effectively determine the scope of the quantifier. Hence the name of $\SI$,
short for Scope Island. If we want to model clause boundaries as scope
islands, we can do so by inserting $\SI$ in the lexical entries of clause
constructors (in our case, the verbs).

\negSpaceBeforeAlign

\begin{align*}
  \sem{\abs{loves}} &:= \lam{O S}{\ap{\SI}{(\app{\sem{\abs{loves}}}{O}{S})}} \\
  \sem{\abs{said}_{\abs{is}}} &:= \lam{C S}{\ap{\SI}{(\app{\sem{\abs{said}_{\abs{is}}}}{C}{S})}} \\
  \sem{\abs{said}_{\abs{ds}}} &:= \lam{C S}{\ap{\SI}{(\app{\sem{\abs{said}_{\abs{ds}}}}{C}{S})}}
\end{align*}

Whenever we use the semantic brackets on the right-hand side of these
revised definitions, they stand for the denotations we have assigned
previously.

\begin{exe}
  \ex Every man loves a woman.
  \ex John said every woman loves me.
  \ex John said, ``Every woman loves me''.
\end{exe}

\negSpaceBetweenAlign

\begin{align}
  & \sem{\app{\abs{loves}}{(\ap{\abs{a}}{\abs{woman}})}{(\ap{\abs{every}}{\abs{man}})}} \nonumber \\
  & \tto \etaE{(\forall x. \ap{\obj{man}}{x} \to (\exists y. \ap{\obj{woman}}{y} \land \app{\obj{love}}{x}{y}))} \\
  & \app{\withSpeaker}{s}{\sem{\app{\abs{said}_{\abs{is}}}{(\app{\abs{loves}}{\abs{me}}{(\ap{\abs{every}}{\abs{woman}})})}{\abs{John}}}} \nonumber \\
  & \tto \etaE{(\app{\obj{say}}{\obj{j}}{(\forall x. \ap{\obj{woman}}{x} \to \app{\obj{love}}{x}{s})})} \\
  & \sem{\app{\abs{said}_{\abs{ds}}}{(\app{\abs{loves}}{\abs{me}}{(\ap{\abs{every}}{\abs{woman}})})}{\abs{John}}} \nonumber \\
  & \tto \etaE{(\app{\obj{say}}{\obj{j}}{(\forall x. \ap{\obj{woman}}{x} \to \app{\obj{love}}{x}{\obj{j}})})}
\end{align}

The calculus offers us flexibility when modelling the semantics. We might
choose to relax the constraint that clauses are scope islands by keeping
the old entries for verbs that do not use the $\SI$ handler. We might then
want to add the $\SI$ handler to the lexical entry of
$\abs{said}_{\abs{ds}}$, next to the $\withSpeaker$ handler, so that
quantifiers cannot escape quoted expressions. We might also allow for
inverse scope readings by, e.g., providing entries for transitive verbs
that evaluate their arguments right-to-left (though then we would have to
watch out for crossover effects if we were to add anaphora).

\subsection{Conventional Implicature}
\label{ssec:ci}

Our goal is to show the modularity of this approach and so we will continue
and plug in one more phenomenon into our growing fragment: conventional
implicatures, as analyzed by Potts~\cite{potts2005logic}. Specifically, we
will focus on nominal appositives.

\negSpaceBeforeAlign

\begin{align*}
  \abs{appos} &: NP \limp NP \limp NP \\
  \abs{best-friend} &: NP \limp NP
\end{align*}

\negSpaceBetweenAlign

\begin{align*}
  \lex{appos}{\lam{X Y}{X \hsbind (\lam{x}{\ap{\SI}{(\etaE{x}
          \mathbin{\overline{=}} Y)} \hsbind
        (\lam{i}{\app{\op{implicate}}{i}{(\lam{z}{\etaE{x}})}})})}} \\
  \lex{best-friend}{\lam{X}{\obj{best-friend} \apr X}}
\end{align*}

In the denotation of the nominal appositive construction, $\abs{appos}$, we
first evaluate the head noun phrase $X : \sem{NP}$ to find its referent $x
: \iota$. We then want to implicate that $x$ is equal to the referent of
$Y$. The term $\etaE{x} \mathbin{\overline{=}} Y$ (note the line over $=$)
is the term that computes that referent and gives us the proposition we
want. We also want to state that no quantifier from within the appositive
$Y$ should escape into the matrix clause and so we wrap this computation in
the $\SI$ handler to establish a scope island. Finally, we pass this
proposition as an argument to $\op{implicate}$ and we return $x$ as the
referent of the noun phrase.

The point of the $\op{implicate}$ operation is to smuggle non-at-issue
content outside the scope of logical operators. The contribution of an
appositive should survive, e.g., logical negation.\footnote{In our limited
  fragment, we will only see it sneak out of a quantifier.} The place where
we will accommodate the implicated truth-conditions will be determined by
the use of the following handler:

\negSpaceBeforeAlign

\begin{align*}
  \accommodate &: \FF_{\{\typedop{implicate}{o}{1}\} \uplus E}(o) \to \FF_E(o) \\
  \accommodate &= \lam{M}{\ap{\banana{\onto{\op{implicate}}{(\lam{i
            k}{\etaE{i} \dand \ap{k}{\star}})}}}{M}}
\end{align*}

We want conventional implicatures to project out of the common logical
operators. However, when we consider direct quotes, we would not like to
attribute the implicature made by the quotee to the quoter. We can
implement this by inserting the $\accommodate$ handler into the lexical
entry for direct speech.

\negSpaceBeforeAlign

$$
  \sem{\abs{said}_{\abs{ds}}} := \lam{C S}{\ap{\SI}{(S \hsbind (\lam{s}{\ap{\obj{say}}{s} \apr (\app{\withSpeaker}{s}{(\ap{\accommodate}{C})})}))}}
$$

Consider the following three examples.

\begin{exe}
  \ex John, my best friend, loves every woman. \label{ex:ci-from-quant}
  \ex Mary, everyone's best friend, loves John. \label{ex:ci-with-quant}
  \ex A man said, ``My best friend, Mary, loves me''. \label{ex:ci-in-quotes}
\end{exe}

In~\eqref{ex:ci-from-quant}, the conventional implicature that John is the
speaker's best friend projects from the scope of the quantifier. On the
other hand, in~\eqref{ex:ci-in-quotes}, the implicature does not project
from the quoted clause and so it is not misattributed.

\negSpaceBeforeAlign

\begin{align}
  & \app{\withSpeaker}{s}{(\ap{\accommodate}{\sem{\app{\abs{loves}}{(\ap{\abs{every}}{\abs{woman}})}{(\app{\abs{appos}}{\abs{John}}{(\ap{\abs{best-friend}}{\abs{me}})})}}})} \nonumber \\
  & \tto \etaE{((\obj{j} = \ap{\obj{best-friend}}{s}) \land (\forall x. \ap{\obj{woman}}{x} \to \app{\obj{love}}{\obj{j}}{x}))} \\
  & \ap{\accommodate}{\sem{\app{\abs{loves}}{\abs{John}}{(\app{\abs{appos}}{\abs{Mary}}{(\ap{\abs{best-friend}}{\abs{everyone}})})}}} \nonumber \\
  & \tto \etaE{((\forall x. \obj{m} = \ap{\obj{best-friend}}{x}) \land (\app{\obj{love}}{\obj{m}}{\obj{j}}))} \\
  & \sem{\app{\abs{said}_{\abs{ds}}}{(\app{\abs{loves}}{\abs{me}}{(\app{\abs{appos}}{(\ap{\abs{best-friend}}{\abs{me}})}{\abs{Mary}})})}{(\ap{\abs{a}}{\abs{man}})}} \nonumber \\
  & \tto \etaE{(\exists x. \ap{\obj{man}}{x} \land \app{\obj{say}}{x}{((\ap{\obj{best-friend}}{x} = \obj{m}) \land (\app{\obj{love}}{(\ap{\obj{best-friend}}{x})}{x}))})}
\end{align}

\subsection{Summary}

Let us look back at the modularity of our approach and count how often
during the incremental development of our fragment we either had to modify
existing denotations or explicitly mention previous effects in new
denotations.

When adding quantification:
\begin{itemize}
\item in the old denotations of verbs, we added the new $\SI$ handler so
  that clauses form scope islands
\end{itemize}

When adding appositives and their conventional implicatures:
\begin{itemize}
\item in the old denotations $\sem{\abs{said}_{\abs{ds}}}$, we added the new
  $\accommodate$ handler to state that conventional implicatures should not
  project out of quoted speech
\item in the new denotation $\sem{\abs{appos}}$, we used the old $\SI$
  handler to state that appositives should form scope islands
\end{itemize}

Otherwise, none of the denotations prescribed in our semantic lexicon had
to be changed. We did not have to type-raise non-quantificational NP
constructors like $\sem{\abs{John}}$, $\sem{\abs{me}}$ or
$\sem{\abs{best-friend}}$. With the exception of direct speech, we did not
have to modify any existing denotations to enable us to collect
conventional implicatures from different subconstituents.

Furthermore, all of the modifications we have performed to existing
denotations are additions of handlers for new effects. This gives us a
strong guarantee that all of the old results are conserved, since applying
a handler to a computation which does not use the operations being handled
changes nothing.

The goal of our calculus is to enable the creation of semantic lexicons
with a high degree of separation of concerns. In this section, we have seen
how it can be done for one particular fragment.

\section{Properties of the Calculus}
\label{sec:properties}

The calculus defined in Sect.~\ref{sec:definition} and to which we will
refer as $\banana{\lambda}$, has some satisfying properties.

First of all, the reduction rules preserve types of terms (subject
reduction). The reduction relation itself is confluent and, for well-typed
terms, it is also terminating. This means that typed $\banana{\lambda}$ is
strongly normalizing.

The proof of subject reduction is a mechanical proof by induction. For
confluence and termination, we employ very similar strategies: we make use
of general results and show how they apply to our calculus. Due to space
limitations, we will pursue in detail only the proof of confluence.

Our reduction relation is given as a set of rules which map redexes
matching some pattern into contracta built up out of the redexes' free
variables. However, our language also features binding, and so some of the
rules are conditioned on whether or not certain variables occur freely in
parts of the redex. Fortunately, such rewriting systems have been
thoroughly studied. Klop's Combinatory Reduction Systems
(CRSs)~\cite{klop1993combinatory} is one class of such rewriting systems.

We will make use of the result that orthogonal CRSs are
confluent~\cite{klop1993combinatory}. A~CRS is \emph{orthogonal} if it is
left-linear and non-ambiguous. We will need to adapt our formulation of the
reduction rules so that they form a CRS and we will need to check whether
we satisfy left-linearity and non-ambiguity (we will see what these two
properties mean when we get to them).

We refer the reader to~\cite{klop1993combinatory} for the definition of
CRSs. The key point is that in a CRS, the free variables which appear in
the left-hand side of a rewrite rule, called metavariables, are explicitly
annotated with the set of all free variables that are allowed to occur
within a term which would instantiate them. This allows us to encode all of
our $x \notin FV(M)$ constraints.

One detail which must be taken care of is the set notation
$(\onto{\op{op}_i}{M_i})_{i \in I}$ and the indices $I$ used in the
$\banana{}$ rules. We do away with this notation by adding a separate rule
for every possible instantiation of the schema. This means that for each
sequence of distinct operation symbols $\op{op}_1$,\ldots,$\op{op}_n$, we
end up with:
\begin{itemize}
\item a special rewriting rule
  $\ap{\banana{\onto{\op{op}_1}{M_1},\ldots,\onto{\op{op}_n}{M_n},\ 
    \onto{\eta}{M_\eta}}}{(\etaE{N})}
  \to \ap{M_\eta}{N}$
\item for every $1 \le i \le n$, a special rewriting rule \\
  $\ap{\banana{\onto{\op{op}_1}{M_1},\ldots,\onto{\op{op}_n}{M_n},\ 
    \onto{\eta}{M_\eta}}}{(\app{\op{op}_i}{N_{\mathrm{p}}}{(\lam{x}{N_{\mathrm{c}}(x)})})}
  \\ \to
  \app{M_i}{N_{\mathrm{p}}}{(\lam{x}{\ap{\banana{\onto{\op{op}_1}{M_1},\ldots,\onto{\op{op}_n}{M_n},\ \onto{\eta}{M_\eta}}}{N_{\mathrm{c}}(x)}})}$
\item for every $\op{op}' \in \EE \setminus \{\op{op}_i \| 1 \le i \le n\}$, a special
  rewriting rule \\
  $\ap{\banana{\onto{\op{op}_1}{M_1},\ldots,\onto{\op{op}_n}{M_n},\ 
    \onto{\eta}{M_\eta}}}{(\app{\op{op}'}{N_{\mathrm{p}}}{(\lam{x}{N_{\mathrm{c}}(x)})})}
  \\ \to
  \app{\op{op}'}{N_{\mathrm{p}}}{(\lam{x}{\ap{\banana{\onto{\op{op}_1}{M_1},\ldots,\onto{\op{op}_n}{M_n},\ \onto{\eta}{M_\eta}}}{N_{\mathrm{c}}(x)}})}$
\end{itemize}

So now we have a CRS which defines the same reduction relation as the rules
we have shown in~\ref{ssec:reductions}. Next, we verify the two
conditions. Left-linearity states that no left-hand side of any rule
contains multiple occurrences of the same metavariable. By examining our
rules, we find that this is indeed the case.\footnote{Multiple occurrences
  of the same $\op{op}_i$ are alright, since those are not metavariables.}

Non-ambiguity demands that there is no non-trivial overlap between any of
the left-hand sides.\footnote{The definition of (non-trivial) overlap is
  the same one as the one used when defining critical
  pairs. See~\cite{klop1993combinatory} for the precise definition.} In our
CRS, we have overlaps between the $\beta$ and the $\eta$ rules. We split
our CRS into one with just the $\eta$ rule ($\to_\eta$) and one with all
the other rules ($\to_{\banana{\lambda}}$). Now, there is no overlap in
either of these CRSs, so they are both orthogonal and therefore confluent.

We then use the Lemma of Hindley-Rosen~\cite[p. 7]{klop1992term} to show
that the union of $\to_{\banana{\lambda}}$ and $\to_\eta$ is confluent when
$\to_{\banana{\lambda}}$ and $\to_\eta$ are both confluent and commute
together. For that, all that is left to prove is that
$\to_{\banana{\lambda}}$ and $\to_\eta$ commute. Thanks to another result
due to Hindley~\cite[p. 8]{klop1992term}, it is enough to prove that for
all $a$, $b$ and $c$ such that $a \to_{\banana{\lambda}} b$ and $a \to_\eta
c$, we have a $d$ such that $b \tto_\eta d$ and $c \to^=_{\banana{\lambda}}
d$. The proof of this is a straightforward induction on the structure of
$a$.

\section{Conclusion}
\label{sec:conclusion}

In our contribution, we have introduced a new calculus motivated by
modelling detailed semantics and inspired by current work in programming
language theory. Our calculus is an extension of the simply-typed lambda
calculus which is the de facto lingua franca of semanticists. Its purpose
is to facilitate the communication of semantic ideas without depending on
complex programming languages~\cite{marsik2014algebraic,kiselyov2010lambda}
and to do so with a well-defined formal semantics.

We have demonstrated the features of our calculus on several examples
exhibiting phenomena such as deixis, quantification and conventional
implicature. While our calculus still requires us to do some uninteresting
plumbing to be able to correctly connect all the denotations together, we
have seen that the resulting denotations are very generic. We were able to
add new phenomena without having to change much of what we have done before
and the changes we have made arguably corresponded to places where the
different phenomena interact.

Finally, we have also shown that the calculus shares some of the useful
properties of the simply-typed lambda calculus, namely strong
normalization.

In future work, it would be useful to automate some of the routine plumbing
that we have to do in our terms. It will also be important to test the
methodology on larger and more diverse fragments (besides this fragment, we
have also created one combining anaphora, quantification and
presupposition~\cite{marsik2014algebraic}). Last but not least, it would be
interesting to delve deeper into the foundational differences between the
approach used here, the monad transformers used by
Charlow~\cite{charlow2014semantics} and the applicative functors used by
Kiselyov~\cite{kiselyov2015applicative}.

%
%
\bibliographystyle{splncs03}
\bibliography{references}

\end{document}